\documentclass[journal,twoside]{IEEEtran}

\IEEEoverridecommandlockouts                              

\makeatletter
\let\NAT@parse\undefined
\makeatother

\usepackage{graphics} 
\usepackage{epsfig} 
\usepackage{mathptmx} 
\usepackage{times} 
\usepackage{amsmath} 
\usepackage{amssymb}  
\usepackage[T1]{fontenc}
\usepackage{booktabs}
\usepackage{adjustbox}
\usepackage{multirow}
\usepackage{siunitx}
\usepackage{booktabs}
\usepackage{makecell}
\usepackage[dvipsnames, table]{xcolor}
\usepackage{makecell}
\usepackage{balance}
\usepackage{cite}
\usepackage{balance}
\usepackage{xcolor}
\usepackage{marvosym}
\usepackage[pagebackref=true,breaklinks=true,colorlinks,bookmarks=false,citecolor=blue]{hyperref}

\hypersetup{
linkcolor=BrickRed
,citecolor=Green
,filecolor=Mulberry
,urlcolor=NavyBlue
,menucolor=BrickRed
,runcolor=Mulberry
,linkbordercolor=BrickRed
,citebordercolor=Green
,filebordercolor=Mulberry
,urlbordercolor=NavyBlue
,menubordercolor=BrickRed
,runbordercolor=Mulberry
}

\definecolor{changeblue}{RGB}{0, 90, 180}

\title{\LARGE \bf
DeepSoMo: Deformable Object Motion Estimation under Proprioceptive Partial Observability via Predictive Latent Dynamics
}

\author{
Hanwen Zhao, Weicheng Dai, Luoyao Chen, Chen Feng\textsuperscript{*}\\
{\tt\small \url{https://ai4ce.github.io/DeepSoMo}}\vspace{-3mm}
\thanks{All authors are with New York University, Brooklyn, NY 11201, USA.}
\thanks{\textsuperscript{*}Corresponding author. This work is supported by NSF Grant 2024882.}
}

\begin{document}

\maketitle
\thispagestyle{empty}
\pagestyle{empty}

\begin{abstract}
Estimating the motion of deformable bodies in soft robotics from proprioceptive sensing is challenging due to partial observability, as internal observations provide only indirect access to global motion, contact, and shape evolution. In such settings, instantaneous observations are insufficient, and accurate estimation requires temporally consistent latent representations. This paper studies deformable motion estimation under proprioceptive partial observability from a latent dynamics perspective. We argue that a key requirement for long-horizon estimation is \emph{latent temporal consistency}, where latent states remain predictive of future observations. To this end, we present \textbf{DeepSoMo}, a learning-based framework that enforces predictive structure in latent dynamics through a self-supervised latent prediction constraint. We evaluate the approach in a canonical deformable dynamics setting using high-fidelity finite-element simulation as a controlled source of ground truth. Results demonstrate consistent improvements in trajectory estimation, orientation recovery, shape reconstruction, and contact inference over architecture-matched recurrent baselines.
\end{abstract}


\section{Introduction}
State estimation is fundamental for robot control, planning, and interaction, but it is especially challenging for soft and deformable robots. Unlike rigid robots, whose states can often be approximated by low-dimensional pose and joint variables, deformable robotic bodies require estimating coupled motion, contact, and shape evolution under nonlinear deformation. This challenge is amplified under proprioceptive sensing, where embedded sensors observe the body only indirectly and may provide limited access to global pose or external contacts. Such sensing is attractive in planetary, subterranean, cluttered, or visually degraded environments, where external cameras, markers, and global localization may be unreliable. The resulting problem is deformable-body motion estimation under proprioceptive partial observability: global motion, orientation, contact, and deformation history must be inferred from incomplete internal observations.

Soft or compliant spherical robots provide a motivating embodiment for this problem. Spherical robots have long been studied as robust mobility platforms for harsh and unstructured environments because their sealed bodies provide omnidirectional mobility, impact tolerance, environmental isolation, and resistance to overturning \cite{yu2025design,raura2017spherical,https://doi.org/10.1049/csy2.12099}. Most existing spherical robots rely on rigid or hard shells, which simplify modeling and control but limit passive adaptation to uneven terrain. Recent inflatable or pressurized spherical designs suggest that compliance can improve impact absorption and terrain conformity \cite{oevermann2024roboball,jangale2025scaling}; however, compliance also makes estimation substantially harder because global motion, contact, and deformation become tightly coupled through nonlinear material response and intermittent terrain interaction.

\begin{figure}
    \centering
    \includegraphics[width=1.0\linewidth]{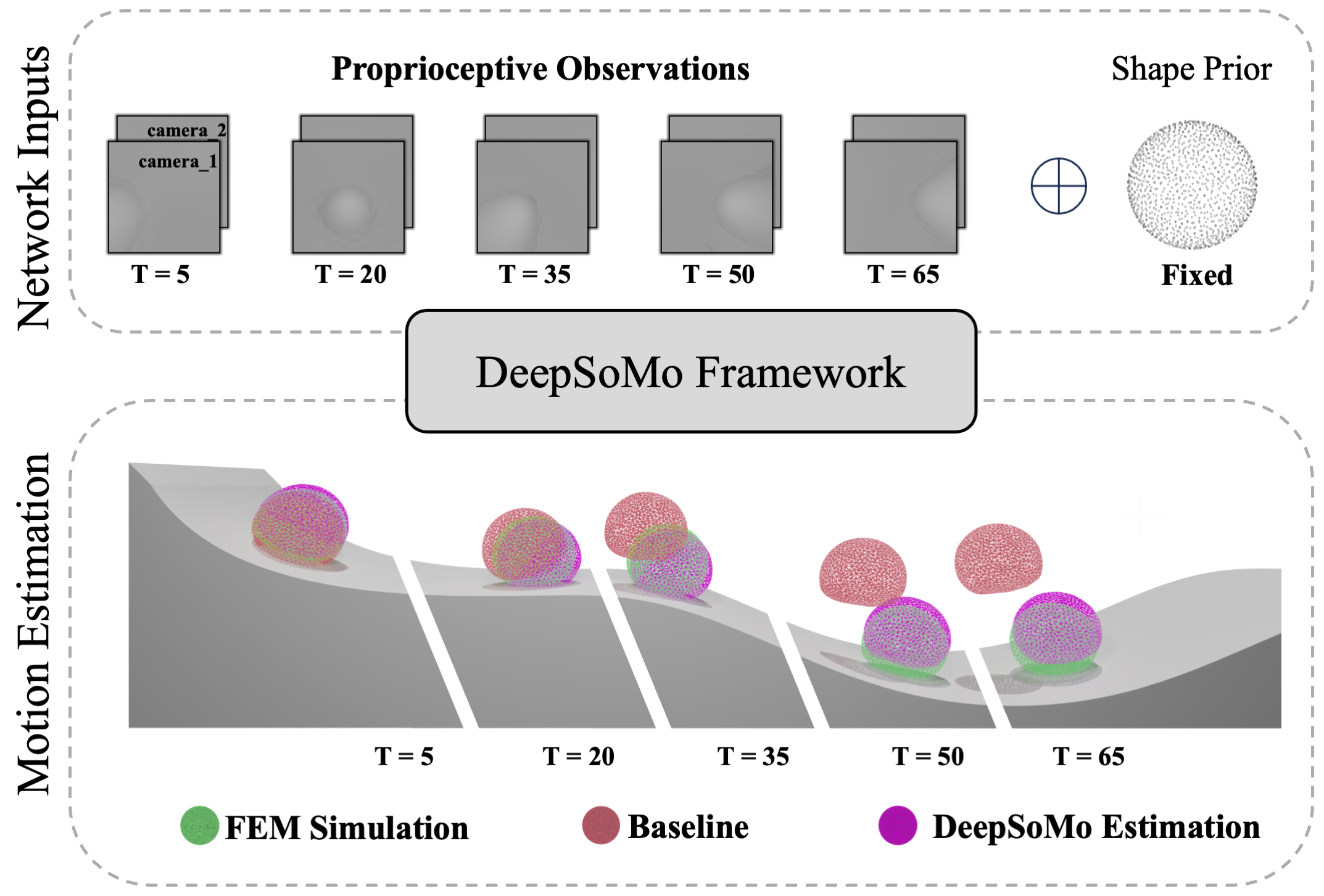}
    \caption{\textbf{DeepSoMo} estimates deformable motion states from proprioceptive visual observations over time. \textbf{Top:} Sequences of internally observed deformation images at different time steps, serving as the only sensory input. A fixed shape prior is used as a reference for deformable shape reconstruction. \textbf{Bottom:} Motion state estimation results in a canonical deformable dynamics scenario, including global pose, shape, and contact evolution. Finite-element-method (FEM) simulation is used solely to generate ground-truth data for evaluation (green). Compared to an architecture-matched recurrent baseline (red), \textbf{DeepSoMo} (purple) produces temporally consistent estimates that remain closer to the ground truth over long horizons, illustrating the effect of enforcing predictive consistency in latent dynamics.}  
    \label{fig:teaser_figure}
\end{figure}


In robotics and machine learning, significant progress has been made in inferring physical dynamics from exteroceptive visual observations, particularly for rigid-body systems \cite{shi2015convolutional,battaglia2013simulation,li2018learning}. Extending such approaches to deformable systems under proprioceptive sensing remains challenging due to high-dimensional state spaces, nonlinear material behavior, and complex contact dynamics.

Recent work in soft robotics has explored learning-based modeling of deformable systems, including particle-based simulators \cite{li2018learning} and vision-based proprioception for recovering high-fidelity \emph{static} deformable shapes using embedded cameras \cite{wang2020real}. While these approaches demonstrate that internal observations can encode rich geometric information, they focus on static or near-static configurations and do not address dynamic motion estimation, where translation, rotation, contact, and shape evolution are tightly coupled over time.

A central difficulty in dynamic deformable motion analysis is \emph{partial observability}. Proprioceptive visual observations reflect only local deformations and provide no direct reference to global pose or external contacts. As a result, instantaneous observations are insufficient to recover the underlying physical state, and accurate estimation requires aggregating information over time. Although recurrent models are commonly used for temporal aggregation \cite{shi2015convolutional}, unconstrained latent dynamics in deformable systems often suffer from error accumulation and latent drift. Prior work in visual dynamics suggests that latent representations optimized only to explain current observations are insufficient for capturing long-term temporal evolution \cite{battaglia2013simulation,chang2016compositional}. These observations motivate the need for latent representations that are explicitly constrained to remain predictive of future observations, which we refer to as \emph{latent temporal consistency}.

This paper studies deformable motion state estimation under proprioceptive partial observability through latent temporal consistency. We argue that enforcing predictive structure in latent dynamics is a key modeling principle for improving observability and robustness in deformable systems, such as soft spherical robots operating in rough or uncertain environments.

As illustrated in Fig.~\ref{fig:teaser_figure}, we evaluate this principle in a canonical deformable dynamics setting inspired by soft spherical robots, where global motion, contact, and shape evolution must be inferred solely from internal visual deformation. The simulated body is used as a controlled testbed for isolating the estimation problem before deployment on physical soft robotic hardware. High-fidelity finite-element simulation generates ground-truth motion and shape data for evaluation. Across this setting, we demonstrate that enforcing latent temporal consistency yields consistent improvements in trajectory estimation, orientation recovery, shape reconstruction, and contact inference compared to architecture-matched baselines.

To address the challenges of deformable motion estimation under proprioceptive partial observability, we present \textbf{DeepSoMo} as a concrete instantiation of the modeling principles outlined above. Our contributions are as follows:

\textbf{(1)} We identify \emph{latent temporal consistency} as a key requirement for estimating the motion states of deformable bodies from proprioceptive visual observations. We show that rich contact and deformation cues captured by internally mounted cameras can support accurate inference of global motion, shape evolution, and contact states when temporal coherence is properly enforced.

\textbf{(2)} We formulate deformable motion analysis as a sequential latent state estimation problem under partial observability, where instantaneous observations are insufficient to recover the underlying physical state. By explicitly modeling temporal dependencies in the latent space, our approach enables long-horizon estimation of coupled translation, rotation, shape, and contact dynamics from proprioceptive inputs.

\textbf{(3)} We introduce a self-supervised latent prediction constraint that regularizes the evolution of latent dynamics by enforcing consistency between current latent states and future observations. This constraint encourages the latent representation to function as a predictive internal state without requiring explicit physical supervision. As a result, the learned latent dynamics exhibit reduced drift and improved long-horizon stability, leading to more robust deformable motion estimation compared to unconstrained recurrent baselines.

\begin{figure*}[tb]
    \label{fig:framework}
    \centering
    \includegraphics[width=0.9\linewidth]{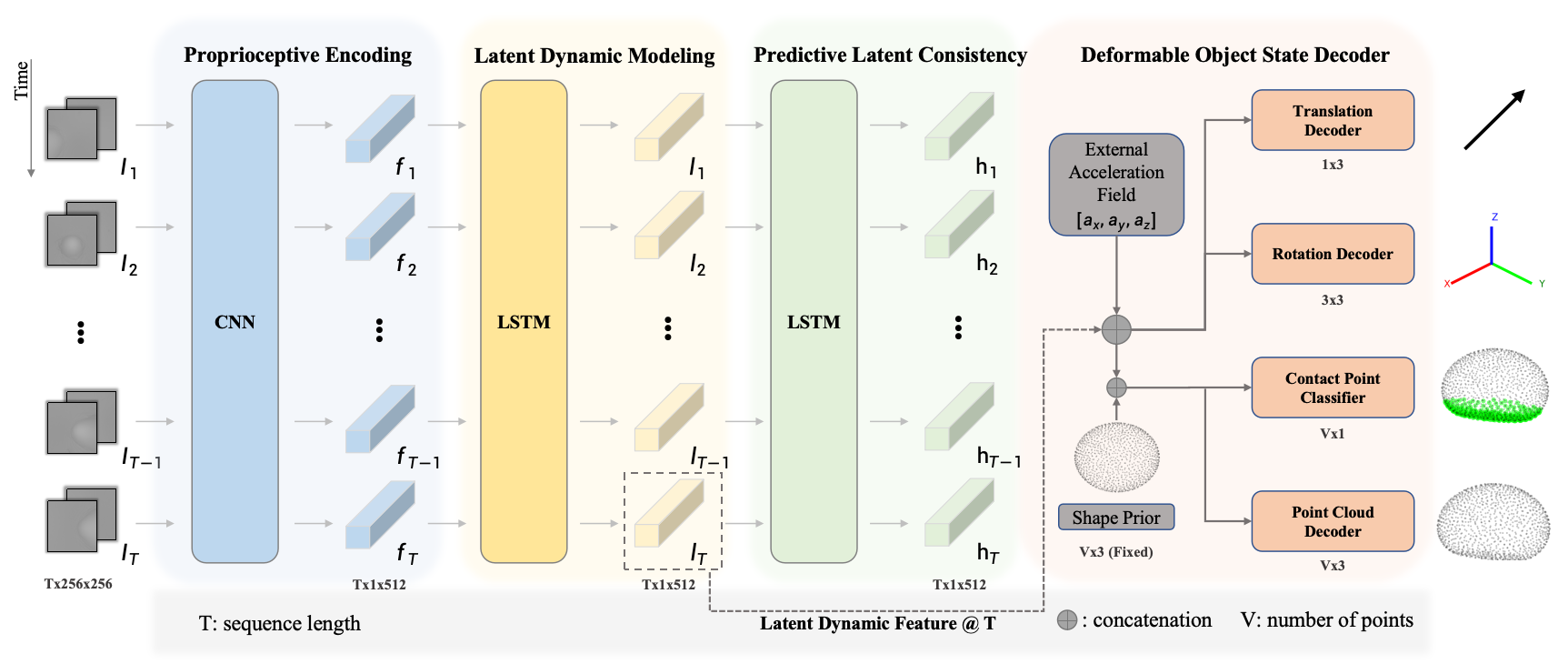}
    \caption{\textbf{Overall framework of DeepSoMo.} 
    Given a sequence of proprioceptive visual observations $I_t$, the framework first encodes each observation into a latent representation $f_t$. A recurrent latent dynamics model aggregates these representations over time to estimate a compact latent state $l_t$ under partial observability. To enforce \emph{latent temporal consistency}, a self-supervised latent prediction module regularizes the latent dynamics by encouraging $l_t$ to remain predictive of future observations, producing a predictive latent state $h_t$. The resulting latent representation is then decoded at each time step into deformable object motion states, including translation, rotation, shape, and contact.}
\label{fig:DeepSoMo}
\end{figure*}


\section{Related Works}

\textbf{Visual Motion Estimation and Physical Reasoning.}
Prior work on visual motion understanding largely falls into two categories: visual prediction and learning physical dynamics from observations. Early data-driven approaches focused on predicting motion events or outcomes from visual inputs, such as nearest-neighbor retrieval for event prediction \cite{yuen2010data} and recurrent models for sequence modeling and visual forecasting \cite{shi2015convolutional, HochSchm97, srivastava2015unsupervised}. While these methods demonstrate that temporal visual observations can encode motion information, they typically emphasize appearance-level prediction rather than explicit state estimation.

A complementary line of work studies learning physical reasoning and dynamics from visual observations. Battaglia et al.~\cite{battaglia2013simulation} introduced intuitive physics models that reason about object stability and interactions, while subsequent works explored estimating physical parameters or latent states underlying Newtonian dynamics from images or videos \cite{brubaker2009estimating, mottaghi2016newtonian, wu2015galileo}. Closest to our setting, Fragkiadaki et al.~\cite{fragkiadaki2015learning} proposed learning predictive models of physical dynamics directly from raw visual input in rigid-body environments, and Ehrhardt et al.~\cite{ehrhardt2019taking} demonstrated that neural networks can implicitly recover physical states from visual data. These approaches focus on rigid-body systems observed through exteroceptive vision and do not address deformable dynamics or proprioceptive partial observability.

\textbf{Soft Robot Proprioception.}
Proprioceptive sensing is widely used in soft robotics to estimate deformation, contact, and interaction forces. Prior work has embedded stretchable or compliant sensors into soft robots to measure bending curvature, force, or tactile information \cite{arriola2020modeling}. Examples include ionogel-based soft sensors integrated via 3D printing \cite{truby2019soft} and internally illuminated elastomer foams for measuring bending and twisting \cite{van2018soft}. While effective for specific designs, these approaches typically rely on low-dimensional representations and are constrained to predefined workspaces.

\textbf{Deformable Object Modeling.}
Modeling deformable objects has long been a challenge in computer graphics and robotics due to their high degrees of freedom and nonlinear material behavior. Classical approaches rely on explicit geometric or physical representations, including implicit surfaces and curves \cite{gascuel1993implicit, cootes1995active, cremers2006dynamical} and parametric spline-based models \cite{catmull1978recursively, prasad2010finding}. Surveys such as \cite{arriola2020modeling} provide comprehensive overviews of these techniques in robotic manipulation.

With the emergence of learning-based methods, discrete representations such as meshes and point clouds have become increasingly popular. Neural architectures including PointNet \cite{qi2017pointnet}, PointNet++ \cite{qi2017pointnet++}, DGCNN \cite{wang2019dynamic}, and FoldingNet \cite{yang2018foldingnet} enable learning directly from point cloud data and have been applied to various 3D perception tasks. Several works have explored learning-based modeling of deformable dynamics, including particle-based interaction networks for deformable simulation and control \cite{li2018learning}. Zhang et al.~\cite{zhang2021deformable} learn locally linear latent dynamics for action-conditioned deformable linear object prediction, while Van der Merwe et al.~\cite{vandermerwe2025estimating} combine learning and model-based optimization to estimate deformable--rigid contact interactions for manipulation. Particularly relevant to our work, DeepSoRo \cite{wang2020real} demonstrated that internal visual observations can be mapped to latent representations that deform a prototype point cloud to recover high-fidelity \emph{static} soft-body shapes. In contrast, our approach addresses proprioceptive dynamic motion estimation, where coupled translation, rotation, shape, and contact states must be inferred over time from internal visual observations under partial observability.

\textbf{Predictive Latent Representation Learning.}
Recent advances in self-supervised representation learning have emphasized learning latent states that are predictive of future observations, rather than reconstructive of the current input. Joint-Embedding Predictive Architectures (JEPA) and related approaches propose learning representations by predicting future latent embeddings, thereby enforcing temporal consistency while avoiding pixel-level reconstruction \cite{assran2023self}. Such methods have demonstrated strong performance in learning temporally coherent and transferable representations from high-dimensional sensory data.

While predictive latent representation learning highlights the importance of temporal consistency, these approaches are developed in the context of generic visual perception and are agnostic to embodiment, contact interactions, and physical state estimation. In contrast, our work applies the principle of predictive latent consistency to deformable motion estimation under proprioceptive sensing, where latent states must encode physically meaningful information about motion, shape, and contact evolution under severe partial observability.


\section{Methods}
\label{sec:methods}

\subsection{Preliminaries}

\textbf{Problem Formulation.}
We study deformable motion state estimation under proprioceptive partial observability. Let $s_t$ denote the deformable object state at time $t$, consisting of body-frame incremental translation $x_t$, body-frame orientation change $r_t$, surface shape $p_t$, and contact state $c_t$. These quantities describe local motion and deformation inferred from internal observations; global pose trajectories can be reconstructed by composing the estimated body-frame motion increments over time.

Due to the proprioceptive sensing setup, the underlying physical state of the system is not directly observable. Instead, the agent receives a sequence of high-dimensional internal visual observations $I_t$ that reflect local deformation and contact. We therefore introduce a latent state $l_t$ that aggregates past observations and serves as an internal representation of the deformable system. Crucially, $l_t$ is not assumed to correspond to a physical state; rather, it is interpreted as a \emph{predictive latent representation} that must remain informative of future observations and motion under partial observability.

Formally, observations are first mapped to latent features
\begin{equation}
    f_t = \phi_{\text{enc}}(I_t),
\end{equation}
which are then aggregated over time via a latent dynamics model
\begin{equation}
    l_t = \phi_{\text{dyn}}(f_0, \dots, f_t),
\end{equation}
and decoded into task-specific local motion and deformation estimates
\begin{equation}
    s_t = \phi_{\text{dec}}(l_t), \quad 
    s_t = \{x_t, r_t, p_t, c_t\}.
\end{equation}

\textbf{Framework Overview.}
As illustrated in Fig.~\ref{fig:DeepSoMo}, DeepSoMo instantiates three modeling components:
(1) a proprioceptive visual encoder that maps internal observations to latent features,
(2) a latent dynamics model that aggregates information over time under partial observability, and
(3) task-specific decoders that map predictive latent states to local deformable motion variables, which can be composed to recover global motion for evaluation.

A self-supervised latent prediction constraint is imposed \emph{during training} to enforce latent temporal consistency, encouraging the shared latent dynamics model to produce states that remain predictive over time.

\begin{figure}[tb]
    \centering
    \includegraphics[width=\linewidth]{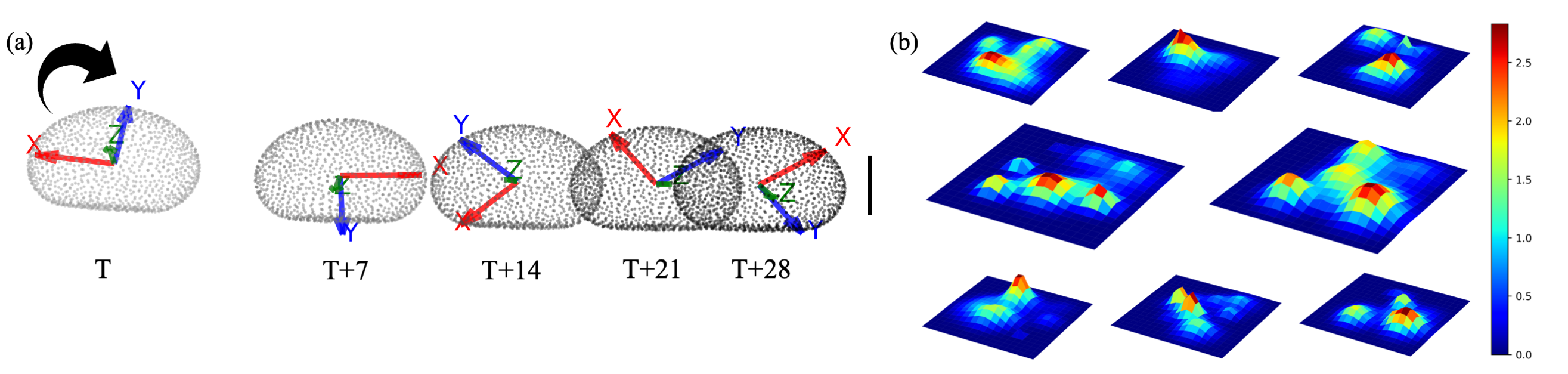}
    \caption{
    (a) Snapshots of a deformable sphere rolling over terrain, illustrating the local body-frame coordinate system used for motion estimation. Translation and orientation are estimated as incremental changes expressed in the body frame attached to the object, and global pose is reconstructed by composing these local estimates over time.
    (b) Examples of procedurally generated terrain surfaces used in simulation. Terrains are generated by adding Gaussian noise to $20 \times 20$m mesh grids, providing controlled variation in elevation and local surface orientation for training and evaluation.}
    \label{fig:dataset}
\end{figure}

\subsection{Dataset Generation}

\textbf{Simulation Platform.}
We use finite-element simulation to generate high-fidelity ground-truth data for supervised training and evaluation. Unlike rigid-body systems, where efficient simulators such as MuJoCo \cite{todorov2012mujoco} or Bullet \cite{coumans2021} are available, deformable dynamics require more computationally intensive methods. In this work, PolyFEM \cite{Li2020IPC} is employed as a data generator due to its robustness under large deformation and computational efficiency. FEM simulation is not used at inference time.

\textbf{Simulation Setup.}
The main dataset consists of a deformable sphere with diameter $0.5\,\mathrm{m}$ rolling over generated terrains. Terrains are created by adding Gaussian noise to $20 \times 20\,\mathrm{m}$ mesh grids, inducing varied contact conditions and temporal deformation-change patterns for learning motion prediction under different object--terrain interactions. To quantify terrain diversity, we compute height-field statistics over the top contact surface of each generated STL mesh. Table~\ref{tab:terrain_stats} summarizes the resulting elevation and slope statistics across the 100 terrain instances. These metrics show that the terrain set contains controlled but meaningful variation in elevation and localized surface orientation. For each of the 100 terrain instances, as illustrated in Fig.~\ref{fig:dataset}, 10 simulation configurations are generated by varying initial pose, terrain scale, and external acceleration fields. Each simulation runs for 100 time steps at $0.1\,\mathrm{s}$ per step, yielding 1000 simulations in total. We reserve 200 simulations from held-out terrain environments for testing and split the remaining 800 into 640 training and 160 validation simulations. \\
For the object-variation experiments, we generate two additional datasets: shape variation and shape--material variation. Each dataset contains 500 simulations, with 100 reserved for testing and the remaining 400 split into 320 training and 80 validation simulations.

\textbf{Proprioceptive Observation Rendering.}
PolyFEM outputs deformable meshes at each time step. Vertex positions are used to form point clouds, from which proprioceptive visual observations are rendered using Blender. Two grayscale cameras are mounted internally, oriented back-to-back, and rigidly attached to the body frame, similar to prior vision-based proprioceptive sensing setups \cite{wang2020real}. Each camera captures a $256 \times 256$ image with a $170^\circ$ field of view. Camera poses are updated according to the body frame, simulating true proprioceptive sensing.

\begin{table}[t]
\centering
\caption{Terrain Diversity Statistics from Generated STL Meshes \\ (mean $\pm$ std over 100 terrain instances).}
\label{tab:terrain_stats}
\begin{adjustbox}{max width=\columnwidth}
\begin{tabular}{l|c}
\toprule
\textbf{Metric} & \textbf{Value} \\
\midrule
Height range (m) & 2.19 $\pm$ 0.59 \\
Height std. (m) & 0.36 $\pm$ 0.09 \\
Mean slope (deg.) & 8.97 $\pm$ 1.57 \\
95th-perc. slope (deg.) & 29.98 $\pm$ 5.26 \\
\bottomrule
\end{tabular}
\end{adjustbox}
\end{table}

\subsection{Model Architecture}

\textbf{Proprioceptive Visual Encoder.}
Each proprioceptive image is processed independently by a shared-weight convolutional encoder based on ResNet-18 \cite{he2016deep}, producing a $512$-dimensional feature vector. Features from the two camera views are concatenated to form the observation embedding $f_t \in \mathbb{R}^{1 \times 512}$.

\textbf{Latent Dynamics Modeling with Self-Supervised Latent Prediction.}
To aggregate information over time, we employ a recurrent latent dynamics model implemented using Long Short-Term Memory (LSTM) units \cite{HochSchm97}. LSTMs are chosen for their stability, simplicity, and well-understood behavior under limited training data. While recurrent aggregation captures temporal dependencies, unconstrained latent dynamics in nonlinear deformable systems often suffer from error accumulation and drift over long horizons \cite{shi2015convolutional}. The proposed predictive latent consistency constraint is agnostic to the specific recurrent architecture and can be applied to alternative sequence models.

\begin{figure}[tb]
    \centering
    \includegraphics[width=0.9\linewidth]{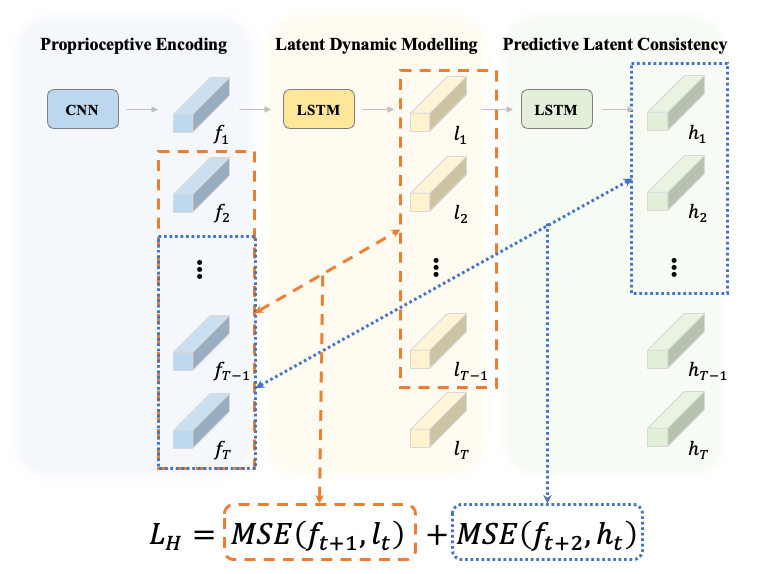}
    \caption{\textbf{Self-supervised latent prediction for enforcing temporal consistency.} 
    Given encoded observation features $\{f_t\}$, the recurrent latent dynamics model produces latent states $\{l_t\}$. To enforce \emph{latent temporal consistency}, a self-supervised prediction constraint encourages the latent state at time $t$ to remain predictive of future observations in the latent feature space. Specifically, $l_t$ is aligned with the next-step encoded feature $f_{t+1}$, and a cascaded predictive latent $h_t$ is further aligned with $f_{t+2}$. This constraint regularizes latent dynamics and mitigates error accumulation over long horizons.}
    \label{fig:hallucination}
\end{figure}

To mitigate this issue, we introduce a \emph{self-supervised latent prediction constraint} that enforces latent temporal consistency as shown in Fig.~\ref{fig:hallucination}. Specifically, the latent state at time $t$ is encouraged to remain predictive of future observations in the latent feature space. Let $l_t$ denote the LSTM output at time $t$. We impose a prediction objective that aligns $l_t$ with the encoded feature at the next time step $f_{t+1}$. This prediction can be further cascaded to produce a higher-order predictive latent $h_t$, similar in spirit to predictive latent representation learning approaches \cite{assran2023self}.

This constraint regularizes the latent dynamics by anchoring them to future observability, rather than relying solely on instantaneous reconstruction or regression losses. Thus, $l_t$ is best understood as a learned predictive state, not an explicit analytical physical state. By aggregating deformation history under partial observability, it is encouraged to retain temporally relevant factors such as rolling direction, accumulated orientation change, contact-induced deformation, and deformation evolution.

\textbf{Local Deformable Motion State Decoders.}
The predictive latent representation is decoded into task-specific local motion and deformation outputs using four separate decoders:

\textit{Body-Frame Translation.}
Due to proprioceptive sensing, global position is not directly observable. The translation decoder estimates incremental translation $x_t$ expressed in a local material body frame. As illustrated in Fig.~\ref{fig:dataset}, the frame uses the mesh centroid as its origin and two fixed material reference regions to determine orientation. Since the mesh topology is fixed, these regions remain attached to the same material locations over time. Each reference region is computed from a small group of nearby vertices for robustness to local deformation artifacts. The resulting reference directions are orthonormalized into a right-handed frame. Global trajectories are reconstructed by composing the predicted body-frame translation and rotation increments over time.

\textit{Body-Frame Orientation Change.}
The orientation decoder estimates the relative orientation change between consecutive time steps, expressed in the body frame. We adopt the continuous 6D rotation representation proposed in \cite{zhou2019continuity}, which avoids discontinuities present in Euler-angle parameterizations and is well suited for gradient-based learning. The decoder outputs this representation, which is converted to a valid rotation matrix via Gram--Schmidt orthogonalization. Global orientation can be recovered by composing the predicted body-frame rotations over time.

\textit{Shape Reconstruction.}
Surface shape is reconstructed using a point-cloud decoder adapted from FoldingNet \cite{yang2018foldingnet}. A fixed prototype point cloud is deformed via point-wise transformations conditioned on the latent state, producing a reconstructed shape in the body frame. Global coordinates can be obtained by transforming the shape using the composed translation and orientation estimates, following approaches presented in \cite{wang2020real}.

\textit{Contact Point Classification.}
A contact classifier predicts whether each point in the reconstructed point cloud is in contact with the environment, enabling explicit modeling of object--environment interaction.

\subsection{Training Objectives}

Ground-truth states generated by FEM simulation are used for supervised training. The overall loss consists of task-specific reconstruction losses and a latent prediction loss.

The latent prediction loss enforces temporal consistency in the latent dynamics:
\begin{equation}
\label{eq:loss_hallu}
L_H = \sum_t \bigl[ \mathrm{MSE}(f_{t+1}, l_t) + \mathrm{MSE}(f_{t+2}, h_t) \bigr].
\end{equation}

The total training loss is given by:
\begin{equation}
\label{eq:loss_total}
L = \sum_t \bigl[ \mathrm{MSE}(x_t) + \mathrm{MSE}(r_t) + \mathrm{CD}(p_t) + \mathrm{BCE}(c_t) \bigr] + L_H,
\end{equation}
where CD denotes the Chamfer Distance \cite{Fan_2017_CVPR} for point-cloud reconstruction and BCE denotes binary cross-entropy for contact classification.

This training objective encourages the latent representation to remain predictive over time, resulting in improved long-horizon deformable motion estimation under proprioceptive partial observability.

\section{Experiments}


\begin{table}[t]
\centering
\caption{Main Results on Deformable Motion Estimation \\ (mean $\pm$ std over 3 seeds).}
\label{tab:main_results}
\begin{adjustbox}{max width=\columnwidth}
\begin{tabular}{ll|ccccc}
\toprule
\textbf{Method} &
\textbf{Prediction} &
\textbf{NTE}$\!\downarrow$ &
\textbf{ATE}$\!\downarrow$ &
\textbf{ARE}$\!\downarrow$ &
\textbf{CPC}$\!\uparrow$ &
\textbf{CD}$\!\downarrow$ \\
\midrule
ResLSTM  & NoPred & 8.3 $\pm$ 0.3 & 56.8 $\pm$ 1.1 & 19.3 $\pm$ 1.0 & 99.0 & 3.1 \\
ObsPred  & Image  & 7.7 $\pm$ 0.5 & 49.9 $\pm$ 7.2 & 19.0 $\pm$ 0.8 & 99.1 & 2.9 \\
\textbf{DeepSoMo} & \textbf{Latent}
& \textbf{6.6 $\pm$ 0.4} & \textbf{38.1 $\pm$ 3.7} & \textbf{16.8 $\pm$ 0.7} & \textbf{99.2} & \textbf{2.5} \\
\bottomrule
\end{tabular}
\end{adjustbox}
\vspace{0.5mm}
\scriptsize{NTE in \%; ATE in cm; ARE in deg.; CPC in \% accuracy; CD in units of $10^{-1}$\,mm.}
\end{table}

\subsection{Baseline Methods}
\label{sec:baselines}

To evaluate the proposed principle of enforcing \emph{predictive structure in latent dynamics}, we compare DeepSoMo against baselines designed to isolate the effect of the latent prediction constraint under the same deformable-state estimation task. The main baselines share the same proprioceptive visual encoder, LSTM dynamics module, and task decoders whenever possible, so performance differences can be attributed primarily to the prediction objective rather than model capacity or output design.

\textit{ResLSTM (NoPred).}
This is the primary architecture-matched baseline. It feeds encoded proprioceptive features directly into an LSTM for temporal aggregation, followed by the same decoders for translation, rotation, shape, and contact estimation. No explicit predictive constraint is imposed on the latent dynamics. This baseline tests whether predictive latent consistency improves long-horizon estimation beyond standard recurrent aggregation under partial observability. \\

\textit{ObsPred (Image).}
This baseline tests an alternative form of predictive supervision in which future prediction is enforced at the observation level rather than in latent space. Encoded features are decoded back into image space using transposed convolutions and bilinear upsampling, and prediction losses are computed between reconstructed and ground-truth images. This comparison evaluates whether image-level prediction provides the same benefit as latent prediction. However, it substantially increases model complexity, containing approximately $156$M trainable parameters compared with $18$M for DeepSoMo. \\

\textit{DeepSoMo (Latent).} 
Our proposed method enforces predictive structure directly in the latent space by aligning latent states with future encoded observations. This latent prediction constraint regularizes temporal dynamics without introducing image-decoder parameters, enabling a direct comparison with both recurrent aggregation and image-level prediction.

\begin{table}[tb]
\caption{Backbone ablations \\ (mean $\pm$ std over 3 seeds).}
\label{tab:ablation_backbone}
\centering
\scriptsize
\setlength{\tabcolsep}{3.2pt}
\renewcommand{\arraystretch}{1.15}
\begin{adjustbox}{max width=\columnwidth}
\begin{tabular}{lcc|ccccc}
\toprule
\textbf{Group} & \textbf{Vision} & \textbf{Dynamics} &
\textbf{NTE}$\!\downarrow$ &
\textbf{ATE}$\!\downarrow$ &
\textbf{ARE}$\!\downarrow$ &
\textbf{CPC}$\!\uparrow$ &
\textbf{CD}$\!\downarrow$ \\
\midrule
\multirow{3}{*}{Vision}
& ResNet & LSTM & \textbf{8.3 $\pm$ 0.3} & \textbf{56.8 $\pm$ 1.1} & \textbf{19.3 $\pm$ 1.0} & \textbf{99.0} & \textbf{3.1} \\
& VGG    & LSTM & 33.5 $\pm$ 0.2 & 757.5 $\pm$ 6.1 & 102.0 $\pm$ 0.02 & 88.8 & 8.1 \\
& ViT    & LSTM & 31.5 $\pm$ 1.5 & 669.6 $\pm$ 65.0 & 70.2 $\pm$ 13.3 & 92.5 & 11.3 \\
\midrule
\multirow{3}{*}{Dynamics}
& ResNet & RNN         & 22.9 $\pm$ 9.4 & 403.2 $\pm$ 307.8 & 46.5 $\pm$ 9.7 & 98.6 & 6.3 \\
& ResNet & DCC         & 13.8 $\pm$ 0.2 & 145.7 $\pm$ 3.2 & 34.4 $\pm$ 0.5 & \textbf{99.0} & 4.3 \\
& ResNet & Transformer & 79.5 $\pm$ 4.4 & 3592.4 $\pm$ 413.0  & 110.3 $\pm$ 11.4 & 88.8 & 12.8 \\
\bottomrule
\end{tabular}
\end{adjustbox}
\vspace{0.5mm}
\scriptsize{NTE in \%; ATE in cm; ARE in deg.; CPC in \% accuracy; CD in units of $10^{-1}$\,mm.}
\end{table}

\subsection{Evaluation Metrics}
\label{sec:evaluation_metrics}
\emph{Normalized Trajectory Error (NTE)} measures trajectory drift relative to the total ground-truth trajectory length. For a sequence of length $T$, we compute
\[
\mathrm{NTE} =\frac{\frac{1}{T}\sum_{t=1}^{T}\|\hat{\mathbf{p}}_t-\mathbf{p}_t\|_2}{\sum_{t=2}^{T}\|\mathbf{p}_t-\mathbf{p}_{t-1}\|_2}\times 100\%,
\]
where $\hat{\mathbf{p}}_t$ and $\mathbf{p}_t$ denote the reconstructed and ground-truth global positions at time step $t$, respectively. NTE provides a scale-normalized measure of long-horizon drift and is useful because small body-frame rotation errors can accumulate into large absolute trajectory deviations over long sequences.

\emph{Absolute Trajectory Error (ATE)} measures the average Euclidean distance between reconstructed and ground-truth global trajectories. Since all sequences share the same initial pose, no trajectory alignment is performed.

\emph{Absolute Rotation Error (ARE)} reports the angular error between reconstructed global orientations and ground truth.

\emph{Chamfer Distance (CD)} evaluates deformable shape reconstruction accuracy by computing the bidirectional distance between predicted and ground-truth point clouds.

\emph{Contact Point Classification Accuracy (CPC)} reports the percentage of points correctly classified as being in contact or non-contact with the environment.

\subsection{Implementation Details}

All models are implemented in PyTorch~\cite{NEURIPS2019_9015}. Experiments are conducted on a single NVIDIA A100 GPU. Each model is trained for 200 epochs with a batch size of 5 using the Adam optimizer with an initial learning rate of $10^{-3}$ and a ReduceLROnPlateau scheduler. Each experiment is repeated three times with different random seeds to assess robustness. Average training time per model is approximately 5 hours.

During training, losses are computed on body-frame incremental translation and orientation change. For evaluation, global trajectories and orientations are reconstructed by composing these local estimates over time.
For the object-variation experiments, we initialize each corresponding ResLSTM baseline and DeepSoMo model from the checkpoint trained on the main canonical dataset, and then continue training on each variation dataset using the same optimization protocol. For each variation, the models are fine-tuned on the 320 training simulations, selected using the 80 validation simulations, and evaluated on the 100 held-out test simulations.


\begin{figure}[tb]
    \centering
    \includegraphics[width=0.9\linewidth]{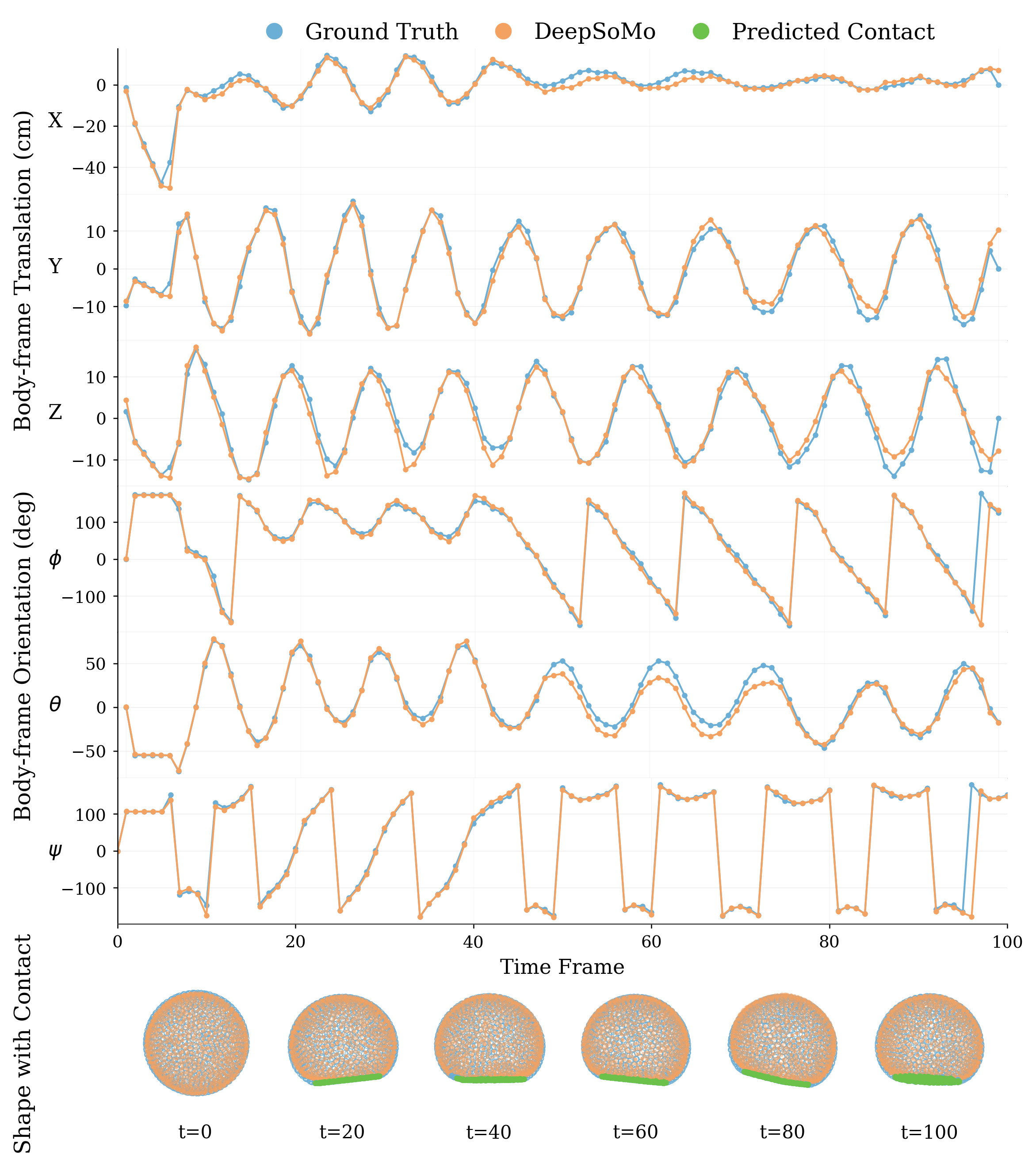}
    \caption{
    \textbf{Qualitative deformable motion estimation under proprioceptive partial observability.}
    Results from a representative unseen test simulation. \emph{Top}: Body-frame incremental translation and orientation estimates over time, comparing ground truth (blue) and DeepSoMo predictions (orange). Translation components are shown as $x$, $y$, and $z$ in centimeters, and orientation components are shown as Euler angles $\phi$, $\theta$, and $\psi$ in degrees. \emph{Bottom}: Deformable shape reconstruction with predicted contact regions (green) overlaid at selected time steps. For this sequence, DeepSoMo achieves an NTE of $4.73\%$, ATE of $23.38\,\mathrm{cm}$, and ARE of $8.84^\circ$, illustrating temporally consistent motion estimation and contact-aware shape reconstruction from internal deformation observations.}
    \label{fig:qualitative}
\end{figure}

\subsection{Performance}
\label{sec:performance}

\textbf{Quantitative Results.}  
Main results are summarized in Table~\ref{tab:main_results}. Across the reported metrics, DeepSoMo consistently outperforms both baselines. DeepSoMo reduces ATE by $32.9\%$ relative to ResLSTM and by $23.6\%$ relative to ObsPred, while also achieving the lowest NTE. These results indicate that predictive latent consistency reduces long-horizon trajectory drift relative to both unconstrained recurrent aggregation and image-level prediction. Because ATE is computed after composing predicted body-frame translation and rotation increments over 100 time steps, it reflects accumulated long-horizon drift rather than frame-wise shape-sensing accuracy. Improvements are also observed in ARE and CD, with comparable CPC across methods.

As shown in Table~\ref{tab:ablation_backbone}, ViT- and Transformer-based variants underperform the ResNet--LSTM backbone despite their larger capacity. We found that their errors are dominated by less stable rotation estimation, which compounds into large global trajectory errors after long-horizon composition. This suggests that, for proprioceptive deformable motion estimation with limited simulation data, locality-preserving visual features and gated recurrent dynamics provide more suitable inductive biases than patch-based visual tokens or full self-attention dynamics.

\textbf{Qualitative Results.}  
Fig.~\ref{fig:qualitative} shows representative qualitative results. Compared to baseline methods, DeepSoMo more accurately tracks global motion, deformation, and contact evolution over time. Additional visualizations and animations are provided in the supplementary material.

\subsection{Ablation Studies and Design Choices}
\label{sec:ablation}

\textbf{Backbone Ablations.}  
Table~\ref{tab:ablation_backbone} reports the impact of varying the vision encoder and latent dynamics backbone. Among vision backbones, ResNet18 significantly outperforms VGG and ViT variants, suggesting that residual architectures provide more stable features for long-horizon latent aggregation. For temporal modeling, LSTM-based dynamics outperform dilated causal convolutions (DCC), vanilla RNNs, and transformers, indicating that gated recurrence remains effective under severe partial observability.

\begin{table}[tb]
\caption{Prediction strategy ablations (mean $\pm$ std over 3 seeds).}
\label{tab:ablation_strategy}
\centering
\scriptsize
\setlength{\tabcolsep}{3.2pt}
\renewcommand{\arraystretch}{1.15}

\begin{adjustbox}{max width=\columnwidth}
\begin{tabular}{lccc|ccccc}
\toprule
\textbf{Group} & \textbf{Prediction} & \textbf{Target} & \textbf{Horizon} &
\textbf{NTE}$\!\downarrow$ &
\textbf{ATE}$\!\downarrow$ &
\textbf{ARE}$\!\downarrow$ &
\textbf{CPC}$\!\uparrow$ &
\textbf{CD}$\!\downarrow$ \\
\midrule
\multirow{2}{*}{Strategy}
& Image  & $f_{t+1}$ & 1
& 7.7 $\pm$ 0.5 & 49.9 $\pm$ 7.2 & 19.0 $\pm$ 0.8 & 99.1 & 2.9 \\
& \textbf{Latent (DeepSoMo)}
& $\boldsymbol{f_{t+1}}$ & 1
& \textbf{6.6 $\pm$ 0.4} & \textbf{38.1 $\pm$ 3.7} & \textbf{16.8 $\pm$ 0.7} & \textbf{99.2} & \textbf{2.5} \\
\midrule
\multirow{2}{*}{Horizon}
& Latent & $f_{t+1}$ & 2
& 6.8 $\pm$ 0.3 & 40.0 $\pm$ 3.6 & 17.7 $\pm$ 1.2 & \textbf{99.2} & 2.9 \\
& Latent & $f_{t+1}$ & 3
& 6.7 $\pm$ 0.1 & 39.0 $\pm$ 0.8 & 17.3 $\pm$ 0.2 & 99.1 & 2.6 \\
\midrule
\multirow{2}{*}{Target}
& Latent & $l_{t+1}$ & 1
& 6.8 $\pm$ 0.1 & 39.7 $\pm$ 2.4 & 17.3 $\pm$ 0.4 & \textbf{99.2} & \textbf{2.5} \\
& Latent & $f_{t+2}$ & 1
& \textbf{6.6 $\pm$ 0.1} & 38.2 $\pm$ 0.7 & 17.5 $\pm$ 0.5 & \textbf{99.2} & 2.6 \\
\bottomrule
\end{tabular}
\end{adjustbox}
\vspace{0.5mm}
\scriptsize{NTE in \%; ATE in cm; ARE in deg.; CPC in \% accuracy; CD in units of $10^{-1}$\,mm.}
\end{table}

\textbf{Prediction Strategy Ablations.}  
Table~\ref{tab:ablation_strategy} analyzes the effect of different prediction strategies. Enforcing prediction in latent space consistently outperforms image-level prediction while requiring far fewer parameters. Increasing prediction horizon beyond one step yields marginal gains, while alternative prediction targets improve over the no-prediction baseline but underperform the proposed design. Overall, aligning latent states with future encoded observations provides the most stable and accurate long-horizon dynamics.

\subsection{Object Variation Experiments}
To evaluate whether predictive latent consistency remains effective beyond the canonical sphere setting, we introduce two controlled object-variation experiments while keeping the sensing setup, terrain generation, and evaluation protocol unchanged.
The canonical object uses a spherical mesh with Neo-Hookean material parameters: Young's modulus $E=1\times10^4$, Poisson's ratio $\nu=0.4$, and density $\rho=500$. The first variation changes only the object shape by anisotropically scaling the same mesh by $[1.6, 1.0, 1.0]$ relative to the canonical object. The second variation changes both shape and material response, using relative scaling $[0.8, 1.0, 1.0]$ and increasing the Young's modulus to $E=3\times10^4$, while keeping $\nu$ and $\rho$ unchanged. These settings test whether the method remains beneficial under altered contact geometry and deformation dynamics. For each variation, ResLSTM and DeepSoMo are initialized from the corresponding canonical checkpoints and fine-tuned on the variation training split. Results are reported in Table~\ref{tab:variation_results}.

\begin{table}[t]
\centering
\caption{Results under Object Variation (mean $\pm$ std over 3 seeds).}
\label{tab:variation_results}
\begin{adjustbox}{max width=\columnwidth}
\begin{tabular}{llccccc}
\toprule
\textbf{Experiment} & \textbf{Method} &
\textbf{NTE}$\!\downarrow$ &
\textbf{ATE}$\!\downarrow$ &
\textbf{ARE}$\!\downarrow$ &
\textbf{CPC}$\!\uparrow$ &
\textbf{CD}$\!\downarrow$ \\
\midrule
\multirow{2}{*}{Canonical sphere}
& ResLSTM  & 8.3 $\pm$ 0.3 & 56.8 $\pm$ 1.1 & 19.3 $\pm$ 1.0 & 99.0 & 3.1 \\
& DeepSoMo & 6.6 $\pm$ 0.4 & 38.1 $\pm$ 3.7 & 16.8 $\pm$ 0.7 & 99.2 & 2.5 \\
\midrule
\multirow{2}{*}{Shape variation}
& ResLSTM  & 7.5 $\pm$ 0.3 & 37.1 $\pm$ 3.4 & 11.4 $\pm$ 0.1 & 98.9 & 4.1 \\
& DeepSoMo & 6.8 $\pm$ 0.8 & 31.5 $\pm$ 2.8 & 11.1 $\pm$ 1.1 & 99.0 & 3.5 \\
\midrule
\multirow{2}{*}{Shape + material}
& ResLSTM  & 9.7 $\pm$ 0.3 & 57.1 $\pm$ 4.3 & 21.5 $\pm$ 0.5 & 99.1 & 3.2 \\
& DeepSoMo & 7.7 $\pm$ 0.6 & 35.6 $\pm$ 6.4 & 19.6 $\pm$ 1.6 & 99.5 & 1.6 \\
\bottomrule
\end{tabular}
\end{adjustbox}
\vspace{0.5mm}
\scriptsize{NTE in \%; ATE in cm; ARE in deg.; CPC in \% accuracy; CD in units of $10^{-1}$\,mm.}
\end{table}

\section{Conclusion}
We presented \textbf{DeepSoMo}, a learning-based framework for deformable motion estimation under proprioceptive partial observability. By leveraging internally observed deformation as a rich proprioceptive signal, DeepSoMo infers coupled motion, shape, and contact states without access to external sensing. The central contribution of this work is the identification and enforcement of \emph{latent temporal consistency} as a key requirement for long-horizon deformable state estimation. We instantiate this principle through a self-supervised latent prediction constraint that regularizes latent dynamics to remain predictive of future observations. Extensive experiments using high-fidelity FEM-generated ground truth demonstrate that this predictive structure substantially reduces latent drift and improves trajectory estimation, orientation recovery, shape reconstruction, and contact inference compared to architecture-matched recurrent baselines.

\textbf{Limitation and Future Work.}
This work is limited by the inherent constraints of proprioceptive visual sensing, which encodes motion only through deformation and contact and therefore cannot observe deformation-sparse motions such as free fall or frictionless sliding. In addition, our experiments focus on simulated single-object interactions with static terrain and do not yet include real-world validation, multi-object interactions, or deformable--deformable collisions. Real-world deployment will require addressing soft-body fabrication, embedded sensing, calibration, and sim-to-real transfer. While the incremental body-frame formulation improves local observability and latent temporal consistency, long-horizon predictions may still accumulate error. Future work will investigate complementary sensing modalities, such as IMUs, tactile or pressure sensors, strain sensors, and actuator feedback, to improve observability, while also extending the framework to physical soft robotic systems and downstream control or planning.


{\small
\bibliographystyle{IEEEtran}
\balance
\bibliography{IEEEabrv,deepsomo}
}


\end{document}